\documentclass[10pt,letterpaper]{article}

\usepackage{cogsci}

 \cogscifinalcopy % Uncomment this line for the final submission 
 \usepackage[table,xcdraw]{xcolor}
\usepackage{graphicx}
\usepackage{amsmath,amssymb}
\usepackage{algorithm}
\usepackage{algorithmic}
\usepackage{pslatex}
\usepackage{apacite}
\usepackage{soul}
\usepackage{subcaption}
\usepackage{float} % Roger Levy added this and changed figure/table
\usepackage[round]{natbib}
\DeclareMathOperator*{\argmax}{argmax}

\newcommand{\E}{\mathbf{E}}
\newcommand{\KL}{\mathbf{KL}}

\title{Learning Approximate and Exact Numeral Systems via Reinforcement Learning}
 
\author{{\large \bf Emil Carlsson (caremil@chalmers.se)} \\
  \AND {\large \bf Devdatt Dubhashi (dubhashi@chalmers.se)} \\
  \AND {\large \bf Fredrik D. Johansson (fredrik.johansson@chalmers.se)} \\
  Department of Computer Science, Chalmers University\\
  Gothenburg, 412 96 Sweden}

\begin{document}

\maketitle

\begin{abstract}
Recent work \citep{Xu2020} has suggested that numeral systems in different languages are shaped by a functional need for efficient communication in an information-theoretic sense. Here we take a learning-theoretic approach and show how efficient communication emerges via reinforcement learning. In our framework, two artificial agents play a Lewis signaling game where the goal is to convey a numeral concept. The agents gradually learn to communicate using reinforcement learning and the resulting numeral systems are shown to be efficient in the information-theoretic framework of \citet{Regier2015,Gibson2016}. They are also shown to be similar to human numeral systems of same type. Our results thus provide a mechanistic explanation via reinforcement learning of the recent results in \citet{Xu2020} and can potentially be generalized to other semantic domains.


\end{abstract}

\section{Introduction}
Why do languages partition mental concepts into words the ways they do? A recent influential body of work suggests language is shaped by a pressure for efficient communication which involves an information-theoretic tradeoff between cognitive load and informativeness \citep{Kemp2012, Gibson2016, Zaslavsky2019a}. This means that language is under pressure to be simultaneously informative, to support effective communication, while also being simple, in order to minimize the cognitive load. 

While the information-theoretic framework is insightful and has broad explanatory power across a variety of domains, see the reviews by \citet{Kemp2018, Gibson2019}, a fundamental question that is left unaddressed is if there is \emph{mechanistic explanation} for how such efficient communication schemes could arise. We address this question here from a learning-theoretic viewpoint: \emph{is there a computational learning mechanism that leads to efficient communication?}

We can relate our approach to previous work using the influential "three levels of analysis" framework posited by David Marr \citep{Marr82} which has been described as one of the most enduring constructs of twentieth century cognitive science and computational neuroscience. While the previous work such as \citet{Kemp2012, Kemp2018,  Gibson2019} is situated at the first or "theory" level of Marr, our analysis is at the \emph{representation and algorithmic} level. In particular, we propose very natural reinforcement learning mechanisms that are able to learn such efficient communication schemes. The learning aspect is emphasised by Tomaso Poggio \citep{Poggio12} in an update of Marr:  
\begin{quote}
  it is ... important to understand how an individual organism, and in fact a whole species, learns and evolves [the computations and the representations used by the brain] from experience of the natural world ... a description of the learning algorithms and their a priori assumptions is deeper, more constructive, and more useful than a description of the details of what is actually learned ... the problem of learning is at the core of the problem of intelligence and of understanding the brain ... learning should be added to the list of levels of understanding ...  
\end{quote}
Recent research gives evidence that the style of learning algorithms we consider here seem to be centrally implicated in exploration strategies used by humans \citep{Schulz2019}.

Reinforcement learning  has been proposed recently as a mechanistic explanation for how efficient communication arises in the colour domain \citep{Kageback2020, Chaabouni} and it was observed that this approach could potentially be applied to other domains. Here we investigate the reinforcement learning approach in the 
domain of numeral systems. It has been shown recently that numeral systems across languages reflect a need for efficient communication \citep{Xu2020}. Numeral systems come in many shapes, some are recursive like English and can express any numeriosity while other non-recursive systems only consists of a small set of words \citep{wals-131}. These non-recursive systems could be either \emph{exact restricted} - in the sense that exact numerosities can only be expressed on a restricted range, or \emph{approximate} like in the language Mundurukú where most numeral words have an imprecise meaning \citep{Pica2004}. Here we only consider non-recursive systems.

We show that reinforcement learning mechanisms can indeed be used to learn exact and approximate numeral systems which are near-optimal in an information-theoretic sense and similar in structure to human numeral systems of the same complexity. Unlike \citet{Kageback2020}, who use a policy-gradient method, we use a Q-learning algorithm with an implicit Thompson Sampling exploration scheme \citep{Sutton1998}.

\section{Learning to Communicate: Signalling Games}
\begin{figure}[t]
\centering
\includegraphics[width=0.3\textwidth]{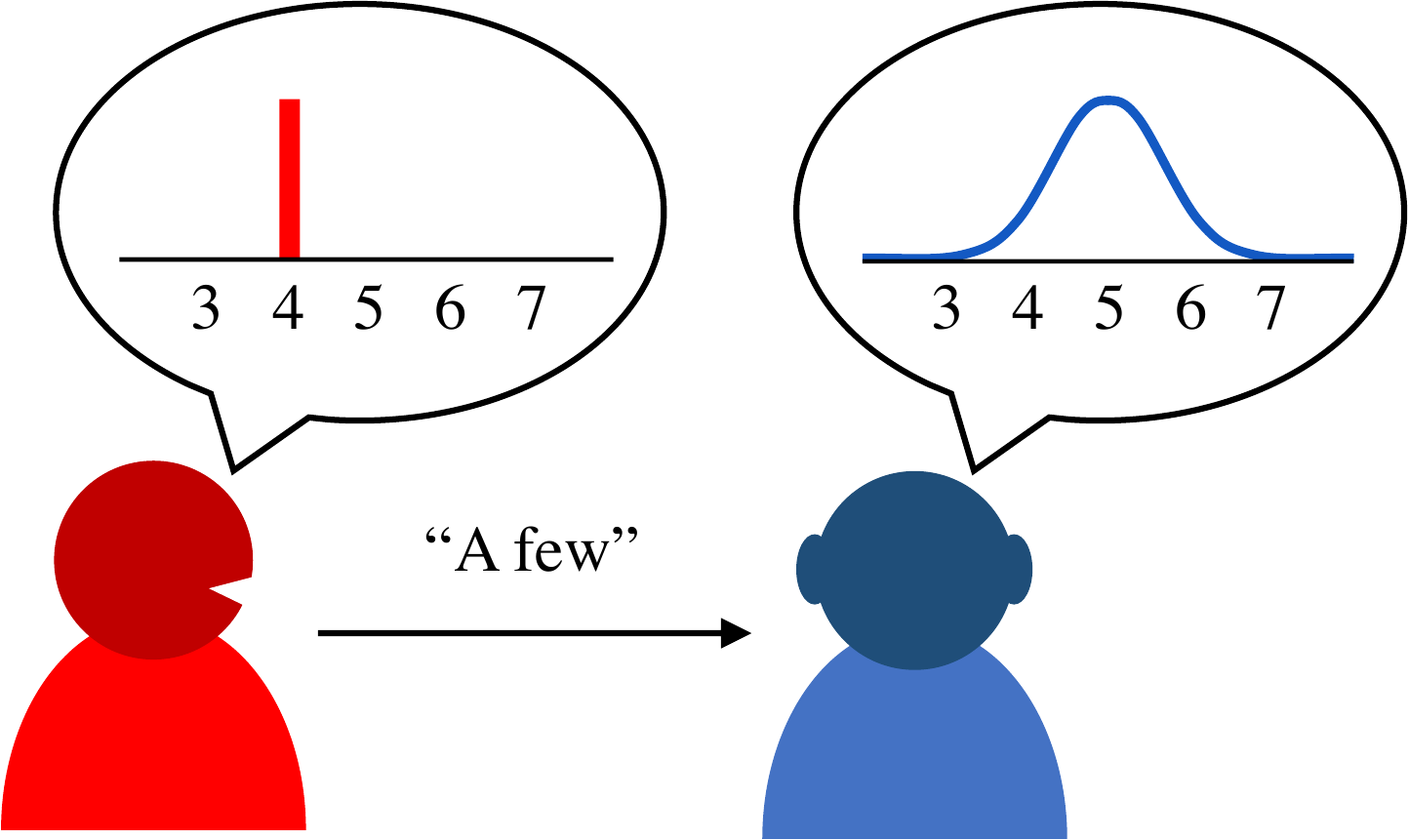}
\caption{Illustration of the communication setup presented in \citet{Xu2020}. The sender wants to convey the numeral concept $4$ and utters ``a few''. The listener is unsure of which numeral the sender is referring to and produces a probability distribution over possible numerals.}\label{fig:regier_model} 
\end{figure}

We consider the communication framework developed in \citet{Regier2015, Xu2020} which consists of a sender and a listener. The sender has a concept in mind and wishes to convey this to a listener over a discrete communication channel. The listener then tries to reconstruct the concept. This is illustrated schematically in Figure \ref{fig:regier_model}.

\begin{figure}[t]
\centering
\includegraphics[width=0.45\textwidth]{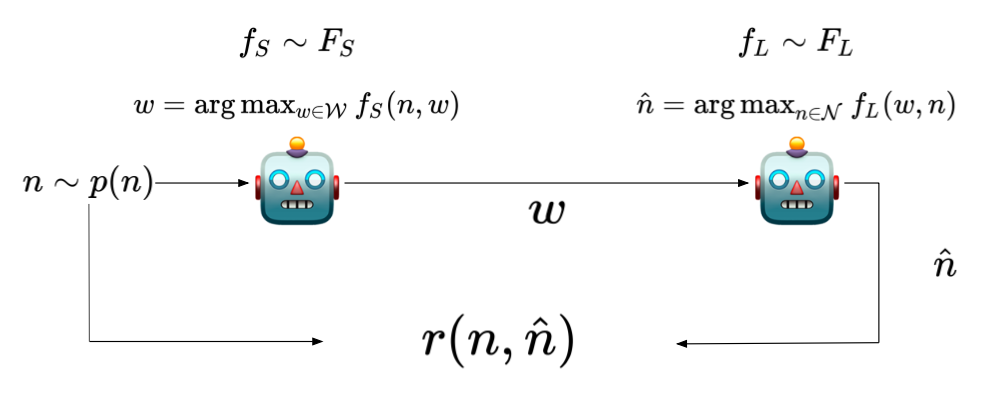}
\caption{Illustration of one round of our Lewis signaling game, which will be formally introduced in later sections. The sender is given a number $n$ and samples a model $f_S$ from $F_S$ using dropout and conveys the word $w$ giving highest reward according to $f_S$. The listener proceeds in similar fashion, given $w$ it samples a model $f_L$ from $F_L$ and guesses the number $\hat{n}$ that yields most reward according to $f_L$. A shared reward is given to both agent based on how close $\hat{n}$ is to $n$.}\label{fig:signaling_game} 
\end{figure}

We extend this setup to a \emph{Lewis signaling game} \citep{Lewis1969}, by considering two artificial agents starting \emph{tabula rasa} and gradually learning to communicate efficiently via a reinforcement learning algorithm (introduced in detail in later sections) by playing several rounds of the game.  In each round of the game, a number $n \in \mathcal{N}$ from the interval $\mathcal{N}$ is sampled according to a need probability of the environment, $p(n)$, which represent how often a numeral concept has to be referred to in the environment. The sampled number $n$ is then given to the sender which has to pick a word $w$ from the vocabulary $\mathcal{W}$ and utter to the listener. Having received a word $w$, the listener guesses a number $\hat{n} \in \mathcal{N}$ and a shared reward, $r(n, \hat{n})$, is given to both agents based on the distance between the guess $\hat{n}$ and the true number $n$. Here we explore three different reward functions, one linear, one inverse and one exponential
\begin{align*}
    &r_{\text{linear}}(n, \hat{n}) = 1 - \frac{|n - \hat{n}|}{|\mathcal{N}|}, \\
    &r_{\text{inverse}}(n, \hat{n}) = (1 + |n - \hat{n}|)^{-1}, \\
    &r_{\text{exp}}(n, \hat{n}) = e^{-|n - \hat{n}|}.
\end{align*}
One round of the signaling game is visualized in Figure \ref{fig:signaling_game} and one could interpret it as follows: the agents are playing a \emph{cooperative game} which involves solving a common task in which success depends on how well the listener reconstructed the number the sender had in mind. The reward functions considered were chosen in order to model different pressure for how precise the listener's reconstruction has to be. % and in Figure \ref{fig:reward_functions} we illustrate of the reward functions behave w.r.t to how close the listener's reconstruction is to the number the sender had in mind. 

\subsection{Reinforcement Learning for Efficient Communication}
Reinforcement learning is an area of machine learning which studies how agents in an environment can learn to pick actions given states as to maximize a reward signal \citep{Sutton1998} and recent studies suggests that reinforcement learning may be an component in neural mechanisms such as the phasic activity of dopamine neurons  \citep{Niv2005, Dabney2020}. In this work our agents will learn to communicate efficiently using reinforcement learning by maximizing the reward in the Lewis signaling game, Figure \ref{fig:signaling_game}. For the sender this translates into conveying the word $w$ which yields highest expected reward given the number $n$ and for the listener to guess the number $\hat{n}$ yielding highest expected reward given the word $w$.

Inherent in this setup is an exploration-exploitation tradeoff---the agents have to balance between exploring uncertain actions in order to gain new insights about the environment and exploiting it current knowledge in order to maximize the reward signal. Recent work in neuroscience suggests that classical machine learning strategies, such as Thompson sampling \citep{Thompson1933}, seem to mechanistically correspond to exploration strategies used by humans \citep{Schulz2019}.

In this work we will use the Bayesian approach and Thompson sampling in order to handle the exploration-exploitation tradeoff. This means that each agent keeps a belief, or posterior distribution, over possible models of the environment and at each time step it samples a plausible model from the belief and acts optimal according to the sampled model. After getting feedback from the real environment an agent updates its belief over possible models accordingly. We will use an implicit form of Thompson sampling presented in \citet{gal2016dropout} where each agent will be represented as a feedforward neural network\footnote{From now on we will use the subscript $S$ for the sender and the subscript $L$ for the listener.} that maps input and action to expected reward 
\begin{align*}
    F_S &: \mathcal{N} \times \mathcal{W} \longrightarrow [0, 1] \\
    F_L &: \mathcal{W} \times \mathcal{N} \longrightarrow [0, 1].
\end{align*}
Given a new round of our signaling game each agent samples a smaller network $f_S \sim F_S$ and $f_L \sim F_L$ from its neural network using the regularization technique dropout \citep{Srivastava2014} which means that the activation at each neuron in the network is randomly set to $0$ with probability $p$. In this way the agents sample, via dropout, one out of all possible models of the expected rewards spanned by $F_S$ and $F_L$. Hence, the networks $f_S$ and $f_L$ become the current internal models of the expected reward of the speaker and listener. Given an input, each agent acts greedily w.r.t. the smaller networks $f_S$ and $f_L$; given the number $n$, the sender conveys the word $\hat{w}$ yielding highest expected reward according the sampled  model \begin{align*}
    \hat{w} = \argmax_{w \in \mathcal{W}} f_S(n, w)
\end{align*}
Similarly, given the word $\hat{w}$, the listener guesses the number $\hat{n}$ satisfying \begin{align*}
    \hat{n} = \argmax_{n' \in \mathcal{N}} f_L(\hat{w}, n').
\end{align*}
 After playing the game for $m$ rounds, each agent update the weights in $F_S$ (or respectively $F_L$) by finding the values which minimize the mean-squared error (MSE) 
\begin{align*}
     \text{MSE}_{S}=\frac{1}{m}\sum_{i}^{m}(f_{S}(\hat{w}_{i}, n_i) - r_i)^2, \\
     \text{MSE}_{L}=\frac{1}{m}\sum_{i}^{m}(f_{L}(\hat{n}_i, \hat{w}_{i}) - r_i)^2.
 \end{align*}
 It should be noted that this game is only partially observable---in each round of the game the sender observes the tuple $(n, \hat{w}, r)$ while the listener observes $(\hat{w}, \hat{n}, r)$.
 
\section{Numeral Systems}
We study two of the three types of numeral systems presented in \citet{Xu2020}. First, we consider the \emph{exact restricted} systems, or simply \emph{exact} systems, where exact numerosities can only be expressed on a restricted range. An example of this is the numeral system \emph{one, two, three} and \emph{more than three}. With this system precise communication can only be achieved for the first three numerals and it is clear which part of the number line each numeral word corresponds to.

The second type is \emph{approximate} numeral systems where the meaning of numerals are approximate. Example of such inexact numerals are \emph{a few} and \emph{many} which do not cover a precise restricted range.

We do not address recursive numeral systems in this work since it require a different way of modelling the agents and we leave it for future work.

\begin{figure*}[t]%
%\centering
\begin{tabular}{ccc}
\multicolumn{3}{c}{\includegraphics[width=0.9\textwidth, height=2cm]{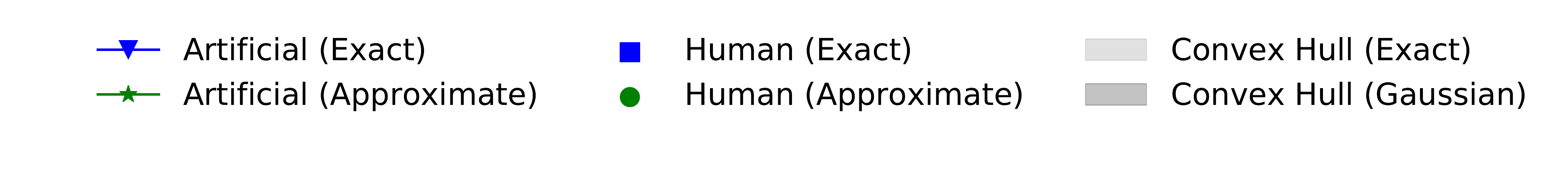}}\\
\subfloat[Reward: Linear,  Prior: Power law]{\includegraphics[width=0.32\textwidth]{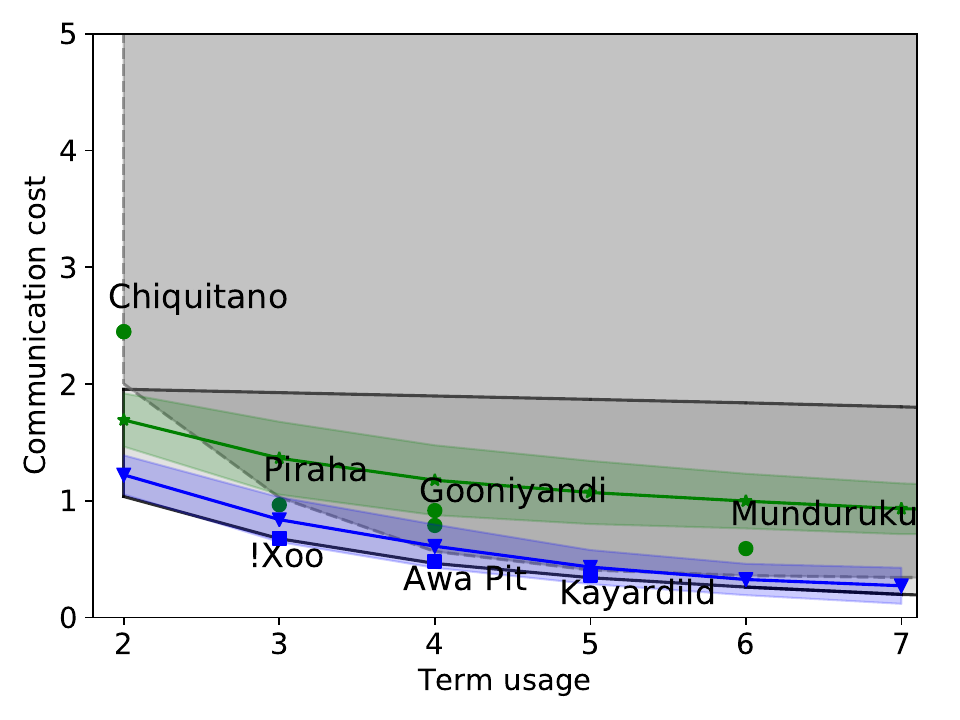}}
 &
  \subfloat[Reward: Inverse ,  Prior: Power law]{\includegraphics[width=0.32\textwidth]{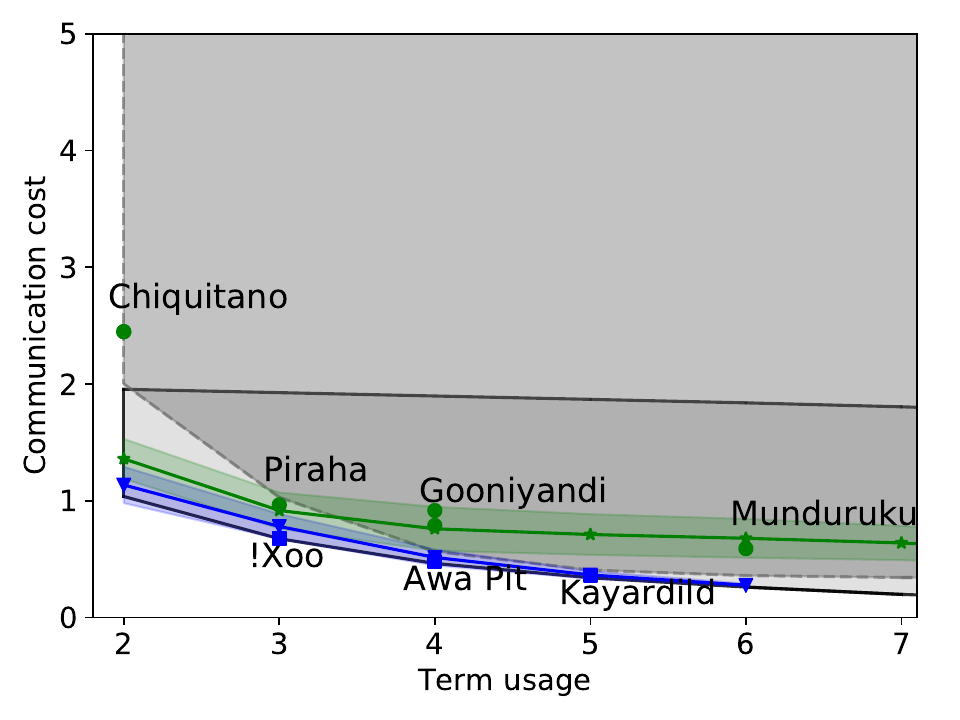}}
  &
 \subfloat[Reward: Exponential,  Prior: Power law]{\includegraphics[width=0.32\textwidth]{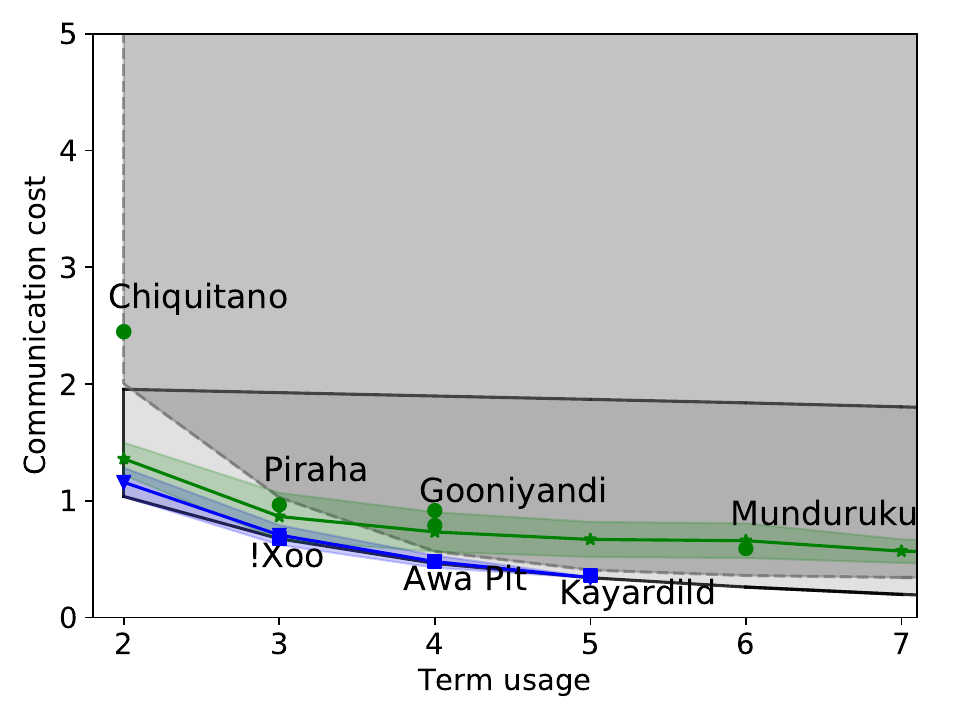}}
\\
\subfloat[Reward: Linear,  Prior: CAP]{\includegraphics[width=0.32\textwidth]{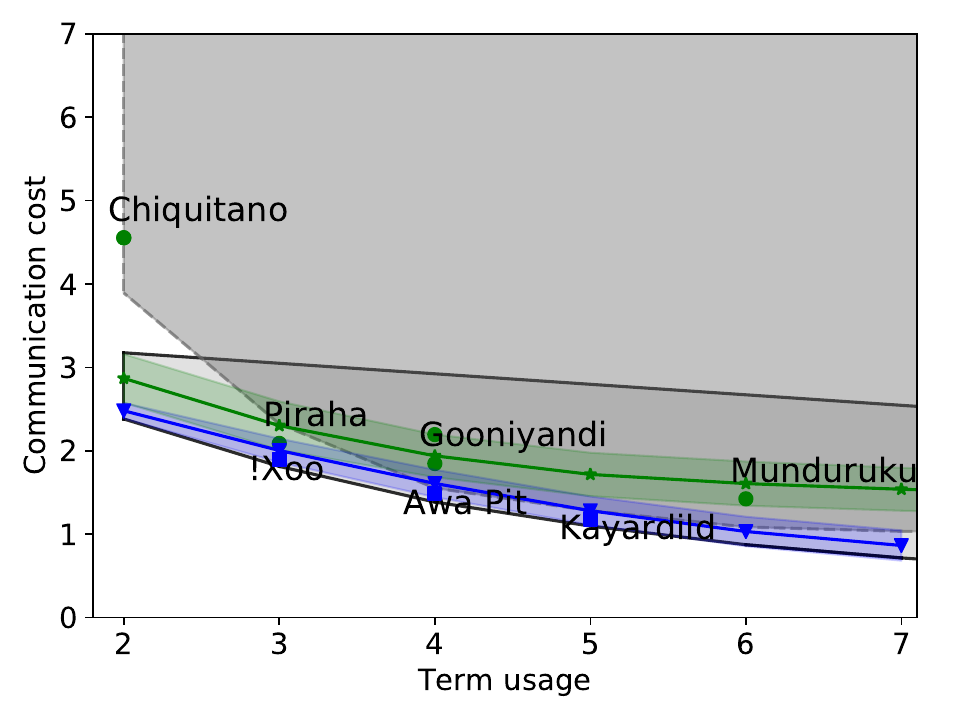}}
&
\subfloat[Reward: Linear,  Prior: MaxEnt]{\includegraphics[width=0.32\textwidth]{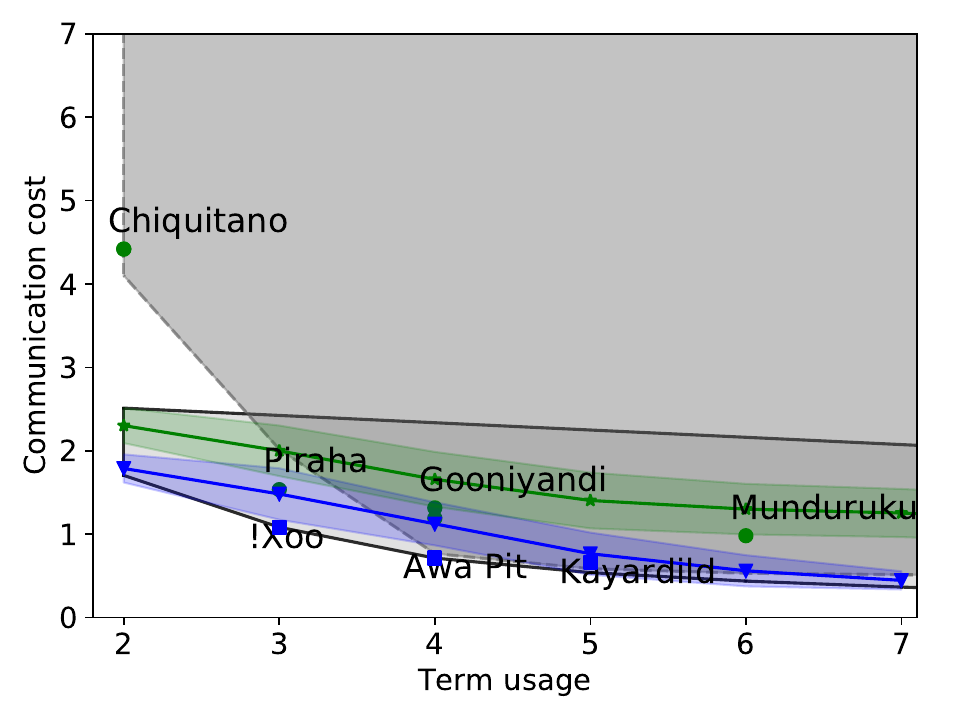}}
&
\subfloat[Reward: Linear,  Prior: Uniform]{\includegraphics[width=0.32\textwidth]{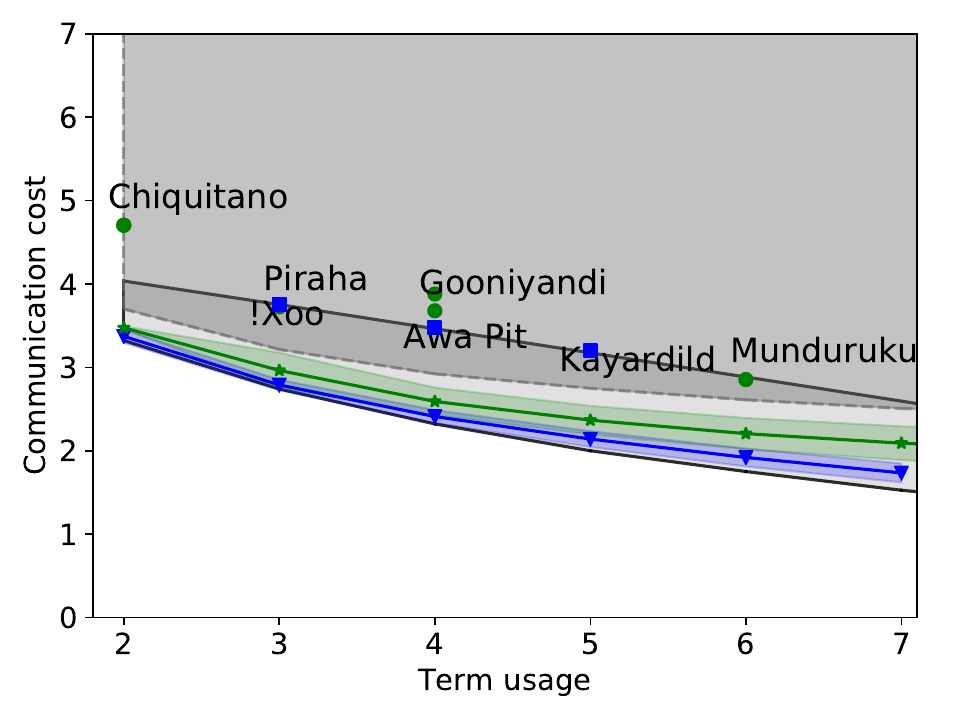}}
\end{tabular}
\caption{Term usage vs communication cost. Note that our agents are not restricted to model the words as Gaussian distributions and can create other probability distributions. This explains why the line goes below the convex hull, for $2$ terms, which was computed assuming Gaussian distributions. We plot the numeral systems from the human languages presented Table \ref{tab:numeral_systems} and since many of them are very similar we only get a few distinct points for human languages in the figure.}\label{fig:convex_hulls}
\end{figure*}
\begin{figure*}
\centering
\begin{tabular}{cccc}
  \includegraphics[width=0.22\textwidth]{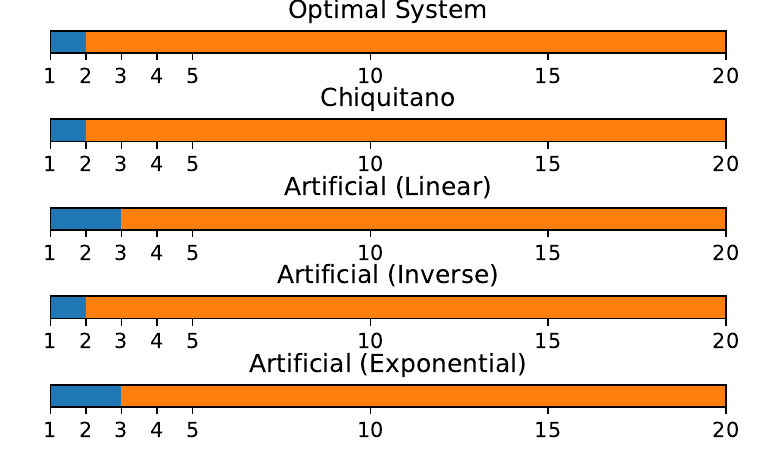}   & \includegraphics[width=0.22\textwidth]{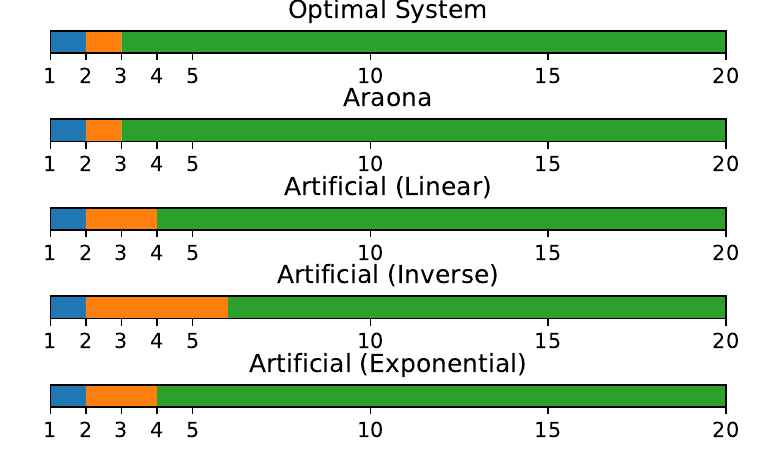} & 
  \includegraphics[width=0.22\textwidth]{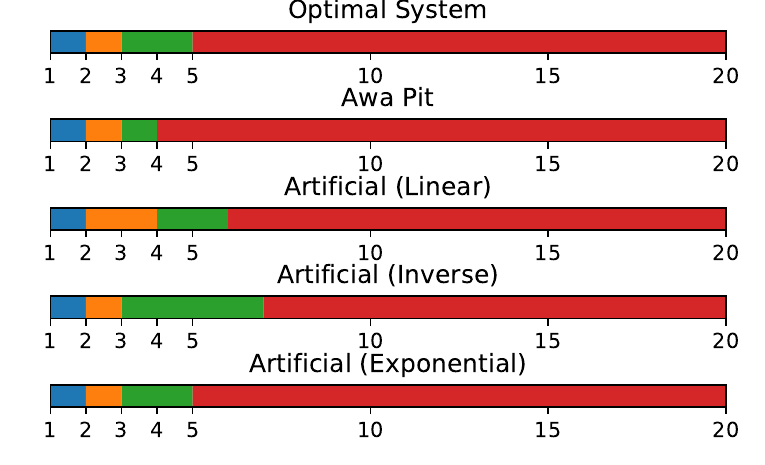} & 
  \includegraphics[width=0.22\textwidth]{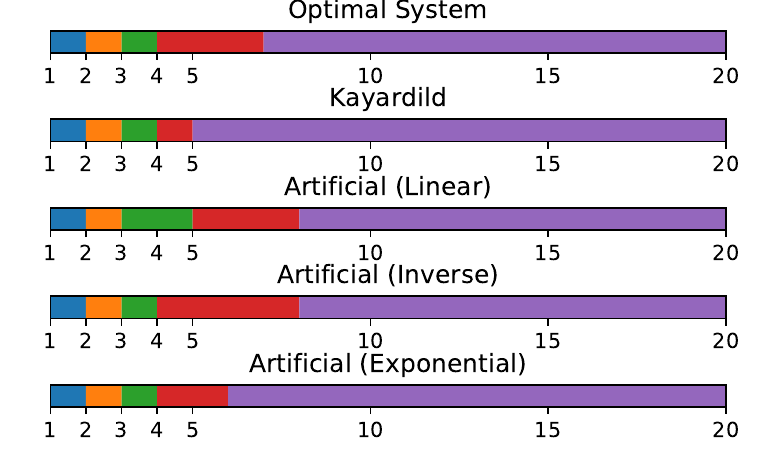} \\
\end{tabular}
\caption{Comparison between the optimal numeral systems w.r.t communication cost, human systems and the artificial consensus systems produced by our agents under the different reward functions. We considered the experiments using the power-law prior and the optimal systems are computed under this prior. Each color represents a numeral word and the corresponding interval on the number line that the word represents.}\label{fig:systems}
\end{figure*}

\subsection{Artificial Numeral Systems}
Given that a sender-listener pair has played the signaling game in Figure \ref{fig:signaling_game} for a certain number of rounds we would like to compute the resulting numeral system. We do this by first estimating the conditional probability $p(w|n)$, i.e the probability that the sender refers to the number $n$ with the word $w$, by running $m=1000$ rounds of the game, without updating the agents, with the number $n$ given to the sender and count how many times each word is used. Hence, we do the following Monte-Carlo estimation  \begin{align*}
     p(w|n) \approx \frac{1}{m} \sum_{i=1}^{m} \mathbf{1}(w = \argmax_{\hat{w}} f_{S, i}(\hat{w}, n))
\end{align*}
where $\mathbf{1}(\cdot)$ is the indicator function. We check if the resulting conditional distribution is peaked, i.e if it for each $n$ assigns more than $0.90$ probability mass to one token $w$, if not we interpret it as an approximate numeral system. Moreover, we consider the mode of $p(w|n)$ to be an exact numeral system. 

\subsection{Complexity and Communication Cost}
We measure complexity of a numeral system simply as the number of words used in the system. In \citet{Xu2020} a grammar based complexity measure was used. This is not needed here since we do not consider recursive numeral systems and for exact and approximate systems there is no pressure for systematicity.

Given a sender distribution $S$ and a listener distribution $L_w$ we measure the communicative cost of conveying a number $n$ as the information lost in the listener's reconstruction of the sender distribution given the numeral $w$. As has been done in previous studies \citep{Xu2020}, we model this as the Kullback-Leibler divergence (KL) between $S$ and $L_w$. Under sender certainty, $S(n)=1$, this reduces to the surprisal \begin{align*}
    \KL(S||L_w) = \sum_{i} S(i) \log \frac{S(i)}{L_w (i)} = - \log L_w (n),
\end{align*}
which can be viewed as how surprised the listener would be by the fact that the sender uttered $w$ if they knew the true number $n$. 

In order to measure the full communication cost of a numeral system we compute the expected surprisal as \begin{align*}
    C = - \sum_{n, w} p(w|n)p(n) \log L_w(n),
\end{align*}
where $L_w(n)$ is computed using Bayes rule\begin{align*}
   L_w(n)  = \frac{p(w|n)p(n)}{\sum_{n^{'}} p(w^|n')p(n')}.
\end{align*} Here $p(w|n)$ denotes the sender partition of the number line and $p(n)$ the need probability of the environment. The measure of the total communication cost of a numeral system used here is exactly the measure of communication cost used in \citet{Gibson2016} and by taking a deterministic sender, i.e a sender which for each $n$ assigns all probability mass to a single word $w$, we get the measure of communication cost used in \citet{Xu2020}. 

Note that we use the speaker model to compute the listener distribution, instead of the listener model, because given a number the sender is forced to assign positive probability to at least one word while the listener can choose to never guess on a number no matter which word is conveyed from the sender. For example the word ``many'' might refer to a large, or possible infinite,  of numbers while the listener may choose to only guess on small subset of these numbers given that ``many'' has been uttered. Another argument for computing the listener distribution using Bayes rule is because, given a sender distribution, it minimizes the communication cost in the information bottleneck framework presented in \citet{Zaslavsky2018a}. The proof of this is presented in the supplementary files of \citet{Zaslavsky2018a}.

\section{Experiments}
\begin{figure}[ht]
    %\centering
    \begin{tabular}{cc}
        \subfloat[\label{fig:need}]{\includegraphics[width=0.23\textwidth]{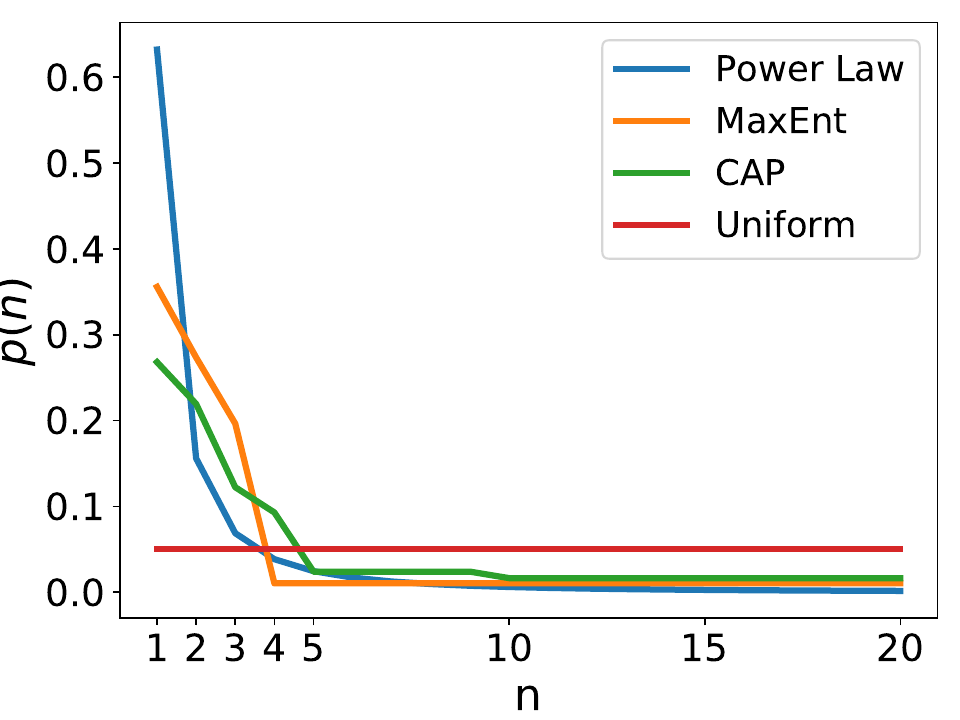}} & \hspace*{-0.5cm}%
         \subfloat[\label{fig:frequency}]{\includegraphics[width=0.23\textwidth]{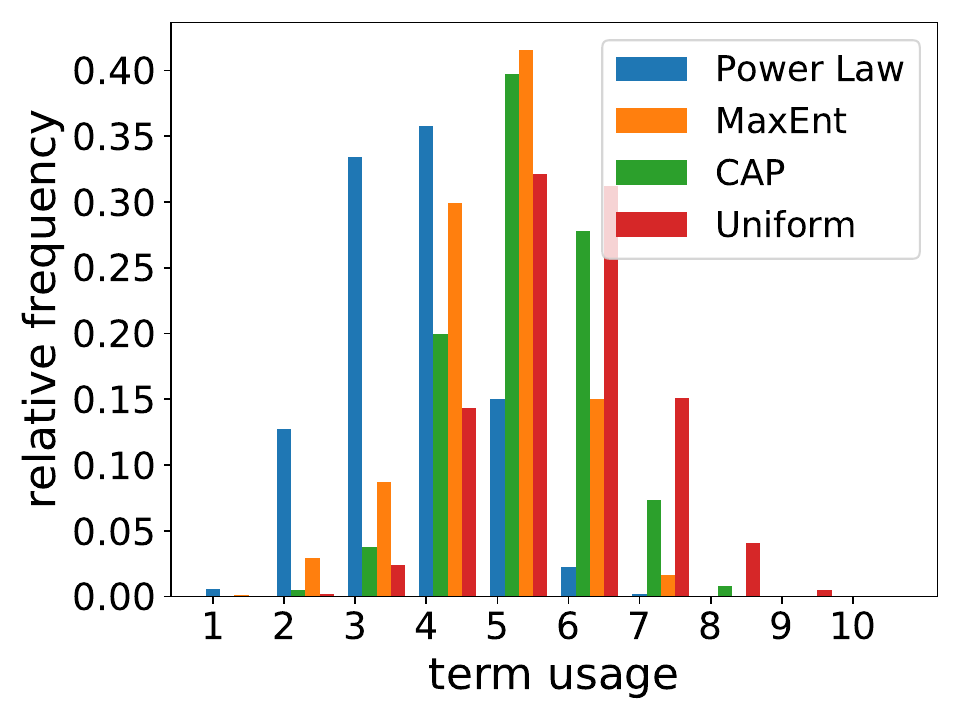}} \\
    \end{tabular}
    \caption{a) The need probabilities, or priors, used. b) Relative frequency of term uses over sender-listener pairs using the linear reward function and varying the need probability. The more left-skewed the need probability is, the fewer terms are generally used by the agents. }
    \label{fig:my_label}
\end{figure}

 We consider the interval $\mathcal{N} = [1, 20]$  and each agent is modelled as a feed-forward neural network with one hidden layer consisting of $50$ hidden neurons with a dropout rate of $p=0.3$ and with ReLu activation \footnote{This interval was chosen since the need distributions are exponentially decaying and very little probability mass lies beyond $20$, see Figure \ref{fig:need}.}. The agents starts with a vocabulary $\mathcal{W}$ \footnote{The size of the vocabulary $\mathcal{W}$ was taken to be equal to the largest number of terms among the human systems analyzed in \citet{Xu2020}, which are presented in Table \ref{tab:numeral_systems}.} of size 10 and is trained for $10\,000$ updates where each update is over a batch of $100$ rounds of the signaling game. The weights in the neural networks are updated using a version of stochastic gradient descent called Adam \citep{Kingma2014} with an initial learning-rate of $0.001$. The dropout rate, learning rate and batch size are in the range of what is commonly used in machine learning. However, we also performed experiments varying these parameters and found the downstream results to be robust. 

We estimate the need probability in four different ways and the priors are shown in Figure \ref{fig:need}. The power-law prior is computed by first taking the normalized frequencies of English numerals in the Google ngram corpus English 2000 \citep{Michel2011} and smoothing the frequencies using a power-law distribution as done in \citet{Xu2020}. We also derive need probabilities using the capacity-achieving prior (CAP) method \citep{Zaslavsky2018a}, which infer a prior directly from naming data, and by using the maximum-entropy (MaxEnt) method \citep{Zaslavsky2019a}, which given a naming distribution $p(w|n)$ and word frequencies $p(w)$ computes the maximum-entropy achieving prior $p(n)$ given these constraints.  We obtain a universal CAP by first computing a CAP for each exact numeral system presented in Table \ref{tab:numeral_systems} and then averaging them together. Further, to compute a MaxEnt prior we consider the language Gooniyandi, which has four number terms translated to \emph{one, two, three, many}, and the corpus data available for the language Gooniyandi \citep[p. 204]{McGregor}. When computing the MaxEnt prior the fourth term, \emph{many}, is modelled as a Gaussian distribution with mean $\mu=5$ and standard deviation $\sigma=0.31 \times \mu$. Lastly, we consider an uniform prior which was also done in \citet{Xu2020} and the authors showed that human systems are less optimal under this prior compare to the more skewed power-law prior, illustrating that the near-optimality patterns found in human numeral systems depend critically on the need probability. 

We start by training $6000$ independent sender-listener pairs under the power-law prior, for each reward function. We then fix the reward function to be linear and train $6000$ independent sender-listener pairs for each of the priors CAP, MaxEnt and Uniform. Note that the agents are free to decide how many terms from the vocabulary that are actually used during communication and it is possible for the agents to converge to a numeral system with less than $10$ terms. Thus, the actual number of terms in the final numeral system will vary over sender-listener pairs due to randomness in the initialization of the neural networks and the sampling from the need probability. 

Following \citet{Xu2020}, we compute the convex hull of hypothetical approximate and exact numeral systems to use as baselines. For exact systems this is done using an approach where we start from a random numeral system and greedily updates the system until a local optima is encountered w.r.t communication cost. For approximate systems we proceed in similar fashion but model a numeral word as Gaussian with a mean $\mu_w$ and a standard deviation $\sigma=0.31 \times \mu_w$ following \citet{Xu2020}. We start from randomly chosen means and perform greedy updates until a local optima is reached. For both types of systems we solve for both the best and worst performing numeral system and the optimization procedures are repeated $1000$ times for each number of terms.

Further, we compare the numeral systems developed by our agents to the human approximate and exact restricted numeral systems considered in \citet{Xu2020} which are presented in Table \ref{tab:numeral_systems}. 
Most of this data was collected from \citet{wals-131} except for Chiquitano, Fuyuge, Krenák which comes from \citet{Harald2010} and Mundurukú which comes \citet{Pica2004}.
\begin{table}[h]
\begin{tabular}{|l|}
\hline
\begin{tabular}[c]{@{}l@{}} \textbf{Approximate systems:}\\ Chiquitano, Fuyuge, Gooniyandi, Mundurukú, Pirahã, Wari\end{tabular}                                                                                                    \\ \hline
    \begin{tabular}[c]{@{}l@{}} \textbf{Exact restricted systems:}\\ Achagua, Araona, Awa Pit, Barasano, Baré, Hixkaryana,\\  Imonda, Kayardild, Krenák, Mangarrayi, \\ Martuthunira, Pitjantjatjara, Rama, Yidiny, !Xóõ\end{tabular} \\ \hline
\end{tabular}
\caption{Human numeral systems considered in Figure \ref{fig:convex_hulls}.}\label{tab:numeral_systems}
\end{table}

In Figure \ref{fig:convex_hulls} we present the performance of our agents, w.r.t communication cost, relative to numeral systems found in human languages and the convex hull of hypothetically possible numeral systems, for the different need probabilities and various reward functions.  We observe that our agents produce numeral systems that are near-optimal for all need probabilities and reward functions. For the left-skewed priors we observe that the communication cost of our agents are close to the communication cost of human systems.

Furthermore, in Figure \ref{fig:frequency} we plot the relative frequency of term usages between the sender-listener pairs when using the linear reward function and varying the need probability. As expected, we observe that a more skewed distribution generally results in fewer terms used by the agents which indicates that numeral systems with few terms can be sufficient to achieve a near-optimal reward while we observe a pressure for using more terms under the uniformed need probability. 

We use Correlation Clustering \citep{Bansal2004} to find the consensus numeral system for each number of terms over all experiments. Correlation Clustering is a method for finding the optimal clustering, w.r.t. a similarity measure. We create a $20 \times 20$ matrix and each time two numbers $i$ and $j$ belongs to the same partition, or word, over two different sender-listener pairs we increase the element $(i,j)$ of the matrix by $1$ otherwise we decrease it with $1$. We apply Correlation Clustering to the final matrix to get a consensus system and this will be an exact numeral system. The resulting systems for the experiments using the power-law prior are presented in Figure \ref{fig:systems} and we observe some similarities between the consensus systems and human systems with the same number of terms. The main difference seems to be that our agents produce systems that tends to be slightly less precise for smaller numbers, especially for the linear reward function, and this could be a result of having reward functions that gives a fair amount of reward for imprecise reconstruction of the number the sender had in mind.

In addition, we compare the representation of numbers developed by our agents to the Gaussian model used in \citet{Xu2020}, which is inspired by the the formalization of the approximate number line presented in \citet{Pica2004}. The model assumes that a numeral word, $w$, is represented as a Gaussian distribution with some mean $\mu_w$ and standard deviation $\sigma=\nu \times \mu_w$ where $\nu$ is the \emph{Weber fraction}. We fit this model to the distributions produced by our agents by first computing, for each sender-listener pair $i$, the expected number $\mu_w^i$ given a word $w$ under the listener distribution $\mu_w^i = \E_{L_w^i}[n|w]$.
We then compute a distribution according to \begin{align*}
    p_{\nu}^{i}(n|w) \propto e^{-(\frac{|n - \mu_w^i|}{2\nu \times \mu_w^i})^2}
\end{align*}
and search for $\nu \in [0.05, 2]$, with a granularity of $0.01$, that minimizes the the MSE w.r.t the listener distribution of pair $i$. The best fitting Weber fractions along with the corresponding MSEs are presented in Table \ref{tab:MSE} and the Gaussian model fits the listener distribution well with an average MSE in the interval $[0.0032, 0076]$ over all the sender-listener pairs. These errors are of the same magnitude as the error reported between the Gaussian model and the numeral system of Mundurukú in \citet{Xu2020} and with similar Weber fraction as reported for Mundurukú adults in \citet{Piazza2013}. Hence, our agents produce approximate numeral systems via reinforcement learning which exhibit similar behavior as the Gaussian models used in \citet{Xu2020} and \citet{Pica2004} without being explicitly programmed to do so. 

\begin{table}[h]
\centering
\begin{tabular}{|l|l|l|}
\hline
\rowcolor[HTML]{C0C0C0} 
Reward      & Best $\nu$ & MSE                 \\ \hline
Linear      & 0.31       & $0.0042 \pm 0.0036$ \\ \hline
Inverse   & 0.31       & $0.0032 \pm 0.0042$ \\ \hline
Exponential & 0.44       & $0.0076 \pm 0.0063$ \\ \hline
\end{tabular}
\caption{The Weber fractions corresponding the Gaussian model that on average fits the listener distribution best along with the average MSE $\pm 1$ standard deviation for that Weber fraction, averaged over all sender-listener pairs trained using the particular reward function. }\label{tab:MSE}
\end{table}

\section{Conclusions and future work}
We have shown that artificial agents can develop exact and approximate numeral systems, via interaction and reinforcement learning, which are near-optimal in an information-theoretic sense and similar to human systems. Our work offers a mechanistic explanation via reinforcement learning of the results in \citet{Xu2020}. More generally, it offers a powerful framework to address fundamental questions of cognition across a wide range of semantic domains using a learning theoretic approach that complements the normative approaches summarized in \citet{Kemp2018, Gibson2019}.

In the numerals domain, there are still several questions that remain to be explored: Would the results be the same if we increase the range of numbers? Can approximate arithmetic be learned in the same way? Could the recursive systems described in \citet{Xu2020} be learned via interaction? An interesting topic for future work is to establish a rigorous connection between reward function and communication cost in our setup. 

In this work our artificial agents have been completely driven by the reward signal. In the future we would like to add a pragmatic reasoning scheme to our model, similar to RSA \citep {Frank12}, and explore what effect this has on the emergent behavior. 
\section{Acknowledgments}
We thank Terry Regier for extremely detailed and valuable feedback on an early draft of this paper. We thank the anonymous reviewers for providing valuable feedback and comments that really improved the final version of the paper. We also thank Meng Liu for clarifying results from \citet{Xu2020}, people at CLASP for providing valuable feedback on an early draft of this paper and Harald Hammarström for valuable comments on an early draft and for providing the reference to the corpus data for Goonyandi. 

This work was supported by funding from CHAIR (Chalmers AI Research Center) and from the Wallenberg AI, Autonomous Systems and Software Program (WASP) funded by the Knut and Alice Wallenberg Foundation. The computations in this work were enabled by resources provided by the Swedish National Infrastructure for Computing (SNIC).

\clearpage
\setlength{\bibleftmargin}{.125in}
\setlength{\bibindent}{-\bibleftmargin}

\bibliography{ref}

\end{document}